\documentclass[9pt, conference]{IEEEtran}
\IEEEoverridecommandlockouts
\usepackage{booktabs}
\usepackage{multirow}
\usepackage{authblk}

\usepackage{cite}
\usepackage{amsmath,amssymb,amsfonts}
\usepackage{algorithmic}
\usepackage{graphicx}
\usepackage{textcomp}
\usepackage{xcolor}
\def\BibTeX{{\rm B\kern-.05em{\sc i\kern-.025em b}\kern-.08em
    T\kern-.1667em\lower.7ex\hbox{E}\kern-.125emX}}
\begin{document}

\title{Towards Child-Inclusive Clinical Video Understanding for Autism Spectrum Disorder}

 \author{Aditya Kommineni$^{1}$, Digbalay Bose$^{1}$, Tiantian Feng$^{1}$, So Hyun Kim$^{2}$, Helen Tager-Flusberg$^{3}$, Somer Bishop$^{4}$, Catherine Lord$^{5}$,\\Sudarsana Kadiri$^{1}$, Shrikanth Narayanan$^{1}$

}

\affil{$^{1}$Signal Analysis and Interpretation Laboratory, University of Southern California, USA\\
$^{2}$School of Psychology, Korea University, Korea\\
$^{3}$Center for Autism Research Excellence, Boston University, Boston, USA\\
$^{4}$Department of Psychiatry, University of California San Francisco, USA\\
$^{5}$Semel Institute of Neuroscience and Human Behavior, University of California Los Angeles, USA\\
}
\maketitle
\begin{abstract}
Clinical videos in the context of Autism Spectrum Disorder are often long-form interactions between children and caregivers/clinical professionals, encompassing complex verbal and non-verbal behaviors. 
Objective analyses of these videos could provide clinicians and researchers with nuanced insights into the behavior of children with Autism Spectrum Disorder. 
Manually coding these videos is a time-consuming task and requires a high level of domain expertise.
Hence, the ability to capture these interactions computationally can augment the manual effort and enable supporting the diagnostic procedure.
In this work, we investigate the use of foundation models across three modalities: speech, video, and text, to analyse child-focused interaction sessions. We propose a unified methodology to combine multiple modalities by using large language models as reasoning agents. We evaluate their performance on two tasks with different information granularity: activity recognition and abnormal behavior detection. 
We find that the proposed multimodal pipeline provides robustness to modality-specific limitations and improves performance on the clinical video analysis compared to unimodal settings.
\end{abstract}


\section{Introduction}
\noindent
Autism Spectrum Disorder (ASD) is a neuro-developmental disorder characterized by challenges with social skills, repetitive behaviors, and nonverbal communication.
According to the Center for Disease Control and Prevention, about 1 in 36 children\cite{maenner2023prevalence} in the United States are diagnosed with ASD, making it one of the leading neuro-developmental disorders in children. 
Currently, diagnosis and behavioral changes during treatment are evaluated through clinically-validated instruments such as Autism Diagnostic Observation Schedule (ADOS) \cite{lord2000autism} and Brief Observation of Social Communication Change (BOSCC) \cite{grzadzinski2016measuring}, which involve dyadic interaction sessions between a clinician/caregiver and the child, involving a multitude of complex activities such as puzzle solving, story creation, toy play and conversation about emotions, loneliness and social difficulties.
\par 
Children on the autism spectrum may exhibit clinically relevant behavior during these sessions such as inconsistent gaze patterns, repetitive behaviors with toys, and repeated words and phrases (echolalia) \cite{lord2020autism, lord2012multisite, lord2018autism}. 
These behaviors are coded either by the clinicians during the assessment or post-assessment by domain experts. 
This annotation process often requires significant manual effort and prevents the scaling of the analysis of these diagnostic sessions.
Hence, the ability to understand and interpret these behaviors computationally could enable clinicians better quantify them, which subsequently allows for automating the analysis of a large number of diagnostic sessions.
\par 
Previously, conversational speech features have been used to identify differences between typically developing children and children on the autism spectrum.  
Vocal entrainment measures \cite{orr1998impact, wynn2018speech, patel2022verbal, lahiri2023context} and prosodic modulations by the clinician \cite{bone2014psychologist} have shown statistically significant correlations to clinical autism scores. 
While these works provide indicators of ASD in speech, they fail to offer a holistic view of the diagnostic session, as they often ignore the dynamics available through the co-occurring visual modality (such as repetitive behaviors, atypical eye gaze, and gesture patterns \cite{ruffman2001social, turner1999annotation, de2020computer}).
\par
Prior works exploring the video modality focus on action recognition \cite{pandey2020guided, sabater2021one}, involving the classification of actions given a video clip of a few seconds, and use facial expressions to ascertain the affective constructs in autism diagnosis \cite{loth2018facial}. 
Additionally, there has been work attempting to characterize autism in children from video clips in the wild \cite{zhang2022discriminative, kojovic2021using, Ali2022VideobasedBU}.
While these works have demonstrated the promise of using video to analyze the behaviors of children with ASD, this area is still under-explored owing to the difficulties in understanding long-form video.
Hence, there is a need for analyzing these sessions in a multimodal manner, which would enable us to better model the observed interaction behaviors with information from both audio and video observation streams.
\par 
\begin{figure*}
    \centering
    \includegraphics[width=0.85\linewidth]{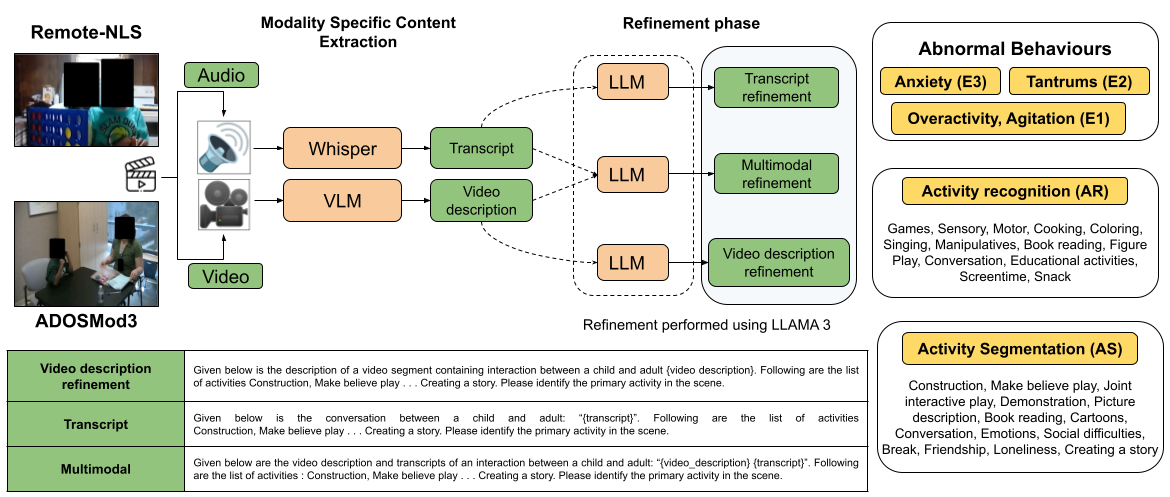}
    \caption{Schematic of the proposed multimodal processing pipeline. During the modality specific content extraction, natural language descriptions of video and speech are obtained. Consequently, these descriptions are used for LLM refinement. E1, E2 and E3 are binary classification tasks. The classes for Activity Recognition and Activity Segmentation are as mentioned. Example prompts corresponding to each refinement mode are provided in the bottom table.}
    \label{fig:enter-label}
\end{figure*}
Recent advances in deep learning have led to impressive abilities in long-form video understanding in domains such as question answering and visual reasoning using foundation models \cite{bommasani2021opportunities}. 
In clinical settings, foundation models \cite{team2023gemini,Saab2024CapabilitiesOG,Yildirim2024MultimodalHA,Nori2023CapabilitiesOG} have shown impressive performance in standard medical examinations and multimodal pathological image analysis. However, the extent to which foundation models can analyze real-world human interactions in a clinical setting is still not well studied.
To the best of our knowledge, this is the first work attempting to explore the utility of these models in the domain of child-adult interactions, especially toward analyzing clinical sessions in the context of ASD. Contributions to this work are as follows:
\begin{itemize}
    \item We propose a unified methodology that combines speech, video and text modalities to achieve robust performance on downstream tasks across different recording settings. 
    \item We provide comprehensive analysis of the capabilities of foundation models to analyze child-adult interactions in autism diagnostic sessions, through evaluating on non-domain (activities) and domain (abnormal behaviors) specific tasks.
    \item  We show noticeable performance gains ($\sim$20\% relative) while reasoning with Large Language Models (LLMs)  through natural language descriptions of video and speech in contrast to zero-shot inference from Video Language Models (VLMs). 
\end{itemize}

\section{Datasets}
\noindent 
For evaluation, we consider two datasets, one containing naturalistic (Remote-NLS) and the other semi-structured clinical (ADOSMod3) child-adult interactions. 
From these datasets, we derive two sets of tasks, i.e., domain-agnostic and domain-relevant tasks. 
Domain-agnostic refers to the set of tasks not related to ASD diagnosis, such as identification of the activities being performed during the video; whereas domain-relevant refers to the set of tasks relevant to ASD diagnosis.
Details about the demographics and recording setting of the datasets are provided in Table.~\ref{tab:datasets}.
We comply with the data usage terms mentioned in the IRB and DUAs from the original data owners.
\subsection{Remote-NLS}
\noindent
\textbf{Remote-NLS} \cite{butler2022remote} contains 89 Zoom recordings of 15-minute child-parent interactions related to
child's spontaneous spoken language in a naturalistic context. An example frame is shown in Fig.~\ref{fig:enter-label} wherein the child is playing a game of Connect4 with their parents. For this dataset, we explore the \textbf{Activity Recognition} task, where the objective is to identify the subset of 13 activities at the session level.
\subsection{ADOSMod3}
\noindent
\textbf{ADOSMod3} comprises child-clinician diagnostic interactions for ASD following the ADOS-2 protocol \cite{lord2000autism, lord2012autism}. The dataset we analyse contains 83 videos of Module 3 ($\sim$1 hr), designed for verbally fluent children. 
Each session is administered to elicit spontaneous interaction and observe verbal and non-verbal communication and behavior of the child through 14 different activities (Fig.~\ref{fig:enter-label}). During the session, clinicians evaluate behaviors associated with communication, interactions, and gesturing by the child.
Unlike Remote-NLS, the ADOS sessions require all pre-defined sets of activities to be performed during the session. Hence, in this scenario, temporal segmentation of activities is important. The task of \textbf{Activity Segmentation (AS)} involves activity prediction at specific time instants. The timestamps of each activity performed were manually annotated for each video.
%\vspace{-6mm}
In addition to information about the activities performed during the course of the session, experts provide codes related to clinically-relevant behaviors exhibited by the children. 
The clinicians conduct this coding process during administering the ADOS session. These \textbf{abnormal behaviors} are coded under three constructs, \textbf{Overactivity/Agitation (E1)}, \textbf{Tantrums (E2)} and \textbf{Anxiety (E3)}. 
For each session, the coding corresponds to binary classification tasks, i.e., presence/absence of behaviors, with low-frequency of abnormal behaviors, the distributions are shown in Table~\ref{tab:results}. The three tasks are defined as below:\\
\textbf{\underline{Overactivity/Aggression (E1):}}
This construct evaluates excessive movement or physical agitation of the child during the ADOS session. This item is coded relative to the participant's nonverbal mental age. Notable characteristics include getting up from the chair, walking around the room during the session and fidgeting or moving about in the chair.\\
\textbf{\underline{Tantrums, Aggression, Disruptive Behavior (E2):}}
This construct refers to any form of anger or disruption displayed by the child during the course of the interaction. This includes behaviors such as occasional mild disruptions in form of anger, aggression, throwing things, hitting or biting others and loud screaming or yelling.\\
\textbf{\underline{Anxiety (E3):}}
For anxiety, clinicians code for signs displayed by the child during the entire session. Notable signs include worry, upset or concern from the child including trembling or jumpiness.
\begin{table*}[]

    \centering
    \caption{Details of the datasets used in the analysis. Age is reported in years, Duration of session is reported in minutes.}
    \resizebox{0.8\textwidth}{!}{
    \begin{tabular}{lcccccccc}
    \toprule
         \textbf{Dataset} & \textbf{Setting} & \textbf{Adult} & \textbf{Duration} & \textbf{\# Sessions} & \textbf{Age} & \textbf{Gender} & \textbf{Tasks} \\
         \midrule
         Remote-NLS & Zoom recordings & Parents & $\sim$15 & 89 & 6.26$\pm$1.07  & 70M, 19F & Activity Recognition\\
         \midrule
         \multirow{2}{*}{ADOSMod3} & \multirow{2}{*}{Diagnostic} & \multirow{2}{*}{Clinician} & \multirow{2}{*}{$\sim$60} &  \multirow{2}{*}{83} & \multirow{2}{*}{8.68$\pm$2.33} & \multirow{2}{*}{56M, 27F} & Activity Segmentation,\\
          &  &  &  &  &  &  & Abnormal behaviors\\
         \bottomrule
    \end{tabular}
    }
    \label{tab:datasets}
\end{table*}
\section{Methodology}
We draw inspiration from works in visual reasoning, robotic planning, and navigation domain \cite{wang2023domino, yang2024empowering, lin2024navcot}, wherein visual entities such as charts and images are converted to natural language through descriptions. Following this, the reasoning agent (LLM) leverages the enriched prompt with these descriptions to perform single-step or multi-step inference for the desired task.
Similarly, rather than performing modality fusion through unified training of vision and speech models, we opt for a training-free alternative wherein task-relevant information from each modality, i.e., video and speech, are extracted from respective pre-trained models (video language models and automatic speech recognition (ASR) models,  respectively). 
Following this, the natural language descriptions for each modality are provided to an LLM to perform inference.
In this manner, the reasoning agent leverages the complementary information present in both modalities without explicitly training a combined model.
Fig.~\ref{fig:enter-label} shows a detailed pipeline for this procedure along with example prompts corresponding to the refinement procedure. 
The textual descriptions of different modalities are integrated into LLM refinement as described below:
\par \noindent 
\textbf{Video} captions are generated by VLMs using a description prompt: 
\begin{verbatim}
    Please provide a detailed description of 
    the video, focusing on the main subjects, 
    their actions, and the background scenes.    
\end{verbatim}
This generated description is provided to an LLM to derive the predictions for the specific task. 
This process allows the VLM to generate a descriptive caption for the video that goes beyond predefined class labels. 
Following this, the LLM is able to infer the downstream task through the contextually rich description.
\par \noindent 
\textbf{Transcriptions} generated from an ASR model (whisper large-v3 \cite{radford2023robust}) are computed for each video. Although the transcriptions would contain errors and would not have the speaker attributions for each utterance, we hypothesize that the information content in the transcripts would contain relevant information for the downstream tasks. Task labels are derived from using the transcriptions corresponding to each segment as prompts to LLMs.
\par \noindent 
\textbf{The multimodal} setting consists of refinement from LLMs by providing both the transcriptions and video captions. The prompt example for this refinement is shown in Fig~\ref{fig:enter-label}. We assume that LLMs could perform better on the tasks by leveraging the complementary information provided by the audio and video streams.
\section{Experiments}
\noindent
Owing to sensitivity surrounding the videos and their clinical context, cloud-based models such as GPT-4 \cite{achiam2023gpt}, Gemini \cite{team2023gemini}, etc., cannot be used. 
Hence, we rely on open-source models which can be deployed in secure local servers.
For video language models, we use LLaVA-NeXT-Video 7B DPO, LLaVA-NeXT-Qwen-32B\cite{zhang2024llavanextvideo} and Video-LLaMA2 7B\cite{cheng2024videollama} models. 
Meta-Llama-3-8B-Instruct\cite{dubey2024llama} is the LLM chosen owing to its large context length and Whisper\cite{radford2023robust} is used for generating the audio transcripts from videos. 
The 7B/8B models are used in 16-bit precision and 32B models are used at 8-bit precision. Single A6000 GPU is used for inference in all experiments. 
\par 
As a baseline, we show the zero-shot performance of the video language models on the respective tasks. 
Then, LLM refinement is performed for three conditions, namely \textit{video only}, \textit{transcript only}, and \textit{multimodal}, as described in the previous section.  
Zero-shot and video descriptions for the video language models are extracted for video segments sampled at 1 fps for 16 seconds. For each video segment, a label is generated for each task.
\par 
Segment-level activity timestamps are not available for activity recognition. Hence, evaluation is performed at the session level. Session labels are determined based on the segment-level activity predictions. A specific activity is recognised to be part of the session if it is predicted by the model for at least 90 seconds.
The threshold is set based on heuristics of the task design. For activity segmentation, evaluation metrics are computed at the segment level as segment-level timestamps are available. Macro F1 scores are reported for both Activity Recognition and Activity Segmentation. 
\par 
For the abnormal activities, we have a single label per session. From the video segment predictions, the ratio of number of segments with presence of abnormal behavior from predictions is computed. Higher ratio indicates more frequent abnormal behavior being observed by the models. This ratio is then used to compute the evaluation metrics.
Since E1, E2 and E3 have significant class imbalance, the Precision Recall Area Under Curve (PR-AUC) metric is reported.
\par 
For \textit{transcript-only} refinement, two settings are evaluated, i.e., transcript chunks corresponding to 16 seconds and 64 seconds video segments. This is done in order to provide a fair comparison for speech modality as 16 seconds of audio may not contain task-specific cues since the transcript would be only 2-3 sentences.
\par 
\section{Results}

\begin{table}
    \centering
    \footnotesize
    \caption{Zero-shot and LLM refinement classification results. Activity \{Recognition, Segmentation\} - AR, AS; overactivity/agitation - E1, tantrums - E2 and anxiety - E3. F1-Macro is reported for AR and AS. PR-AUC is reported for E1, E2 and E3. \textit{A} refers to absence and \textit{P} refers to Presence of the abnormal behavior. For all metrics, higher is better.}
    \resizebox{0.5\textwidth}{!}{
    \begin{tabular}{lccccccc}
    \toprule
    \multirow{2}{*}{Model} & AR & AS & \multirow{2}{*}{E1} & \multirow{2}{*}{E2} & \multirow{2}{*}{E3} \\
     & (CAASL) & (ADOS) & &  \\
    \midrule
    \multirow{3}{*}{\# Classes} & \multirow{3}{*}{13} & \multirow{3}{*}{14} & 2 & 2 & 2 \\
    & & & (\textit{A} - 48) & (\textit{A} - 76) & (\textit{A} - 73)\\
    & & & (\textit{P} - 34) & (\textit{P} - 7) & (\textit{P} - 10)\\
    \midrule
    \textit{Zero-shot} \\
    \quad VidL 2 7B$^a$, & 31.6 & 4.0 & 84.2 & 12.4 & 22.6\\
    \quad L-Next Vid 7B DPO$^b$  & 27.6 & 3.8 & 81.5 & 4.2 & 19.3\\
    \quad L-Next Vid 32B Qwen$^c$ & 36.8 & 4.4 & 91.7 & 54.2 & 28.0\\
    \midrule
    \textit{LLM Refinement} \\ 
    \rotatebox[origin=c]{180}{$\Lsh$}Video only \\
    \quad VidL 2 7B$^a$ & 37.7 & 5.1 & 78.1 & 45.5 & 19.9 \\
    \quad L-Next Vid 7B DPO$^b$ & 36.8 & 4.3 & 76.8 & 54.2 & 23.7 \\
    \quad L-Next Vid 32B Qwen$^c$& 42.8 & 3.7 & 61.3 & 4.2 & 18.2\\
    \rotatebox[origin=c]{180}{$\Lsh$}Transcript only \\
    \quad whisper-L (16s) & 16.3 & 12.3 & 83.9 & 10.7 & 9.8 \\
    \quad whisper-L (64s) & 39 & 22 & 79.1 & 10.4 & 12.0 \\
    \rotatebox[origin=c]{180}{$\Lsh$}Multimodal \\
    \quad VidL 2 7B$^a$ & 41.3 & 12.4 & 80.1 & 37.3 & 16.0 \\
    \quad L-Next Vid 7B DPO$^b$ & 38.7 & 9.6 & 70.0 & 9.4 & 28.8 \\
    \quad L-Next Vid 32B Qwen$^c$ &  44.8 & 9.8 & 82.2 & 10.4 & 12.0 \\
    \bottomrule
    \end{tabular}
    }
    
    \label{tab:results}
    \tiny{$^a$Video-LLaMA2 7B, $^b$LLaVA-NeXT-Video 7B DPO, $^c$LLaVA-NeXT-Qwen-32B}
\end{table}
\newcommand{\zshotres}{\hspace*{.4in}\rotatebox[origin=c]{180}{$\Lsh$}\xspace}
\subsection{Activity Recognition}
\par
\begin{figure}
    \centering
    \includegraphics[scale=0.5]{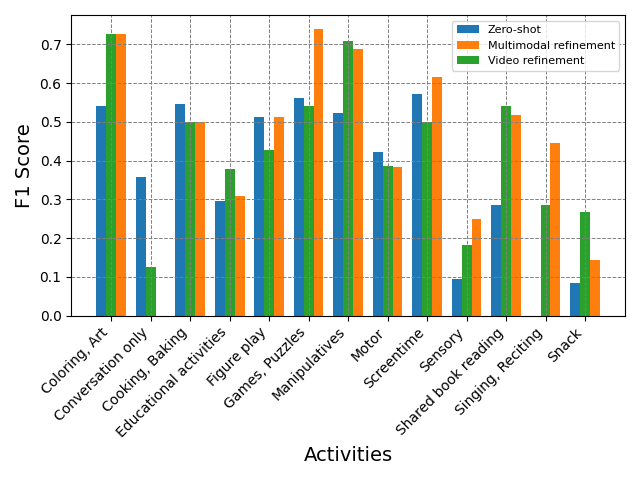}
    \caption{Class wise Activity Recognition F1 for LLaVA-NeXT-Qwen-32B}
    \label{fig:caasl_classwise}
\end{figure}

 \noindent
Table~\ref{tab:results} shows the results for both activities and abnormal behaviors. 
Firstly, for activity recognition, we see that both zero-shot and LLM refinement perform significantly better than random chance. 
Additionally, we see that augmenting the prompt with contextual rich descriptions from videos can improve the LLM refinement performance ($\sim$20\% relative) compared to zero-shot inference from VLMs across all tested models. 
Fig~\ref{fig:caasl_classwise} shows the class-wise F1 scores for activity recognition on the best performing VLM i.e. LLaVA-NeXT-Qwen-32B.
Here, we see that the zero-shot inference is unable to generalize well for nuanced activities such as \textit{manipulatives}, and \textit{sensory}. 
This behavior could be explained by VLMs being better at generating descriptions about visual entities but unable to reason the distinctions between these nuanced categories. 
Hence, LLMs are able to use these descriptive captions of the video to provide better reasoning, thereby better demarcating these classes. Additionally, performance gains are observed with multimodal refinement in tasks such as \textit{singing, reciting} and \textit{shared book reading}, pointing to LLM being able to leverage complementary information from audio and video to perform more accurate reasoning.
\par 
\subsection{Activity Segmentation}
Activity Segmentation on ADOSMod3 results in chance-level performance for zero-shot and video description-only refinement. This could be explained by two factors.
Firstly, the video descriptions generated by the VLMs are unable to capture fine-grained details necessary for the LLM refinement to reason about the downstream tasks. It is worth noting that multiple activities (make-believe play, join interactive play, creating a story) in an ADOS session involve the use of similar toys. The VLM descriptions do not capture these distinctions, thereby causing the LLM refinement to be unable to distinguish these activities. 
Also, ADOSMod3 dataset contains conversational activities (loneliness, emotions, conversation and reporting), wherein the details are solely captured by speech modality. 
Hence, for transcript-only or multimodal LLM refinement, we observe a noticeable performance improvement over video-only LLM refinement, indicating a stronger signal for activity segmentation in speech.
% \par 
\subsection{Abnormal behaviors}
Both zero-shot and LLM refinement are able to capture Overactivity (E1) in the videos as indicated by the results in Table.~\ref{tab:results}. E1 is often characterized by the child getting up from the chair and moving around the room. 
Since VLMs are trained on action recognition and movement tasks, they are able to map the construct of overactivity to movement of the child around the room.  
But in E2, accounting for tantrums and negative behavior requires understanding the voice characteristics (such as loud voice, verbal threats) of the child in addition to the speech content and actions leading to the LLM refinement being unable to generalize well to this task. 
For E3, we observe that the VLM descriptions incorrectly recognize the child as trembling even though the child exhibits regular hand movements as a part of play. This leads to LLM refinement incorrectly predicting the child appearing to be anxious during these video segments.  

\subsection{Multimodality vs Unimodality}
The results for activity segmentation or activity recognition show that some modalities might be more informative about a given task. For example, in the case of activity recognition (CAASL), video modality is more informative than audio. However, in the case of ADOS activity segmentation, we notice the opposite trend. 
The proposed approach of combining modality-specific information followed by LLM refinement enables inferring from the more informative modalities, thereby providing improved performance across the different activity tasks.
However, while we observe performance gains for multimodal refinement in activity tasks, we do not see a similar trend in abnormal behavior tasks. 
As mentioned earlier, this is primarily owing to errors in modality-specific content extraction, i.e., hallucinations in VLMs or ASR models. 
Currently, multimodal refinement equally weighs all the modality descriptions without providing modality-specific confidence weights. This leads to faulty inferences by multimodal refinement in the presence of errors in modality-specific descriptions.

\subsection{Context Scaling and Model Size}
Here we test whether increasing the context length of the input modality results in improved downstream performance. Since most of the video language models considered in this work are trained for a maximum context of 16 frames, we choose to perform this analysis using audio modality, wherein increasing the context length is equivalent to providing transcripts from longer video segments.
\par
As seen from the results in Table~\ref{tab:results}, comparing the rows whisper-L (16s) and whisper-L (64s) indicates that increasing the context length of audio information from 16s to 64s leads to a significant improvement in the performance for activity tasks.
This could be because transcripts which are 16 seconds long may not provide activity-specific information and might have generic conversations which leads to the model falsely predicting those chunks as conversations. When increasing the context length to 64 seconds, there is a higher chance of capturing information about the primary activity.
For the activities task, we see a performance improvement for the 32B llava-next model when compared to the 7B model.
However, no consistent trend is seen for abnormal behaviors. 

\subsection{Hallucinations}
Hallucinations in foundation models is a well-documented phenomenon\cite{bai2024hallucination}. During the inference for both activities and abnormal behaviors, we observe hallucinations which leads to incorrect reasoning. For example in the representative image for Remote-NLS dataset (Fig~\ref{fig:enter-label}), the child is wearing a T-shirt with a basketball logo on it. While describing the video, the models sometimes misidentify the task as the child playing basketball. These hallucinations could potentially be reduced by considering precise object-centered grounding information.
\section{Conclusion \& Future Work}
In this work, we explore the utility of foundation models in analysing clinical videos for ASD. 
The proposed multimodal refinement pipeline provides robustness to unimodal limitations through utilizing complementary information present in the additional modalities. We also underscored the extant limitations of these models in supporting reasoning in these complex human interaction contexts. 
In the future, we plan to expand the work to provide multistep reasoning wherein the LLM reasoning agent would be able to selectively prompt for task specific information. Additionally, we plan to extend the reasoning capabilities to identification of a larger set of ASD related behaviors such repetitive behaviors, gestures and atypical gaze.
\bibliography{main}

\begin{thebibliography}{10}

\bibitem{maenner2023prevalence}
M.~J. Maenner, ``Prevalence and characteristics of autism spectrum disorder among children aged 8 years—autism and developmental disabilities monitoring network, 11 sites, united states, 2020,'' {\em MMWR. Surveillance Summaries}, vol.~72, 2023.

\bibitem{lord2000autism}
C.~Lord, S.~Risi, L.~Lambrecht, E.~H. Cook, B.~L. Leventhal, P.~C. DiLavore, A.~Pickles, and M.~Rutter, ``The autism diagnostic observation schedule—generic: A standard measure of social and communication deficits associated with the spectrum of autism,'' {\em Journal of autism and developmental disorders}, vol.~30, pp.~205--223, 2000.

\bibitem{grzadzinski2016measuring}
R.~Grzadzinski, T.~Carr, C.~Colombi, K.~McGuire, S.~Dufek, A.~Pickles, and C.~Lord, ``Measuring changes in social communication behaviors: preliminary development of the brief observation of social communication change (boscc),'' {\em Journal of autism and developmental disorders}, vol.~46, pp.~2464--2479, 2016.

\bibitem{lord2020autism}
C.~Lord, T.~S. Brugha, T.~Charman, J.~Cusack, G.~Dumas, T.~Frazier, E.~J. Jones, R.~M. Jones, A.~Pickles, M.~W. State, {\em et~al.}, ``Autism spectrum disorder,'' {\em Nature reviews Disease primers}, vol.~6, no.~1, pp.~1--23, 2020.

\bibitem{lord2012multisite}
C.~Lord, E.~Petkova, V.~Hus, W.~Gan, F.~Lu, D.~M. Martin, O.~Ousley, L.~Guy, R.~Bernier, J.~Gerdts, {\em et~al.}, ``A multisite study of the clinical diagnosis of different autism spectrum disorders,'' {\em Archives of general psychiatry}, vol.~69, no.~3, pp.~306--313, 2012.

\bibitem{lord2018autism}
C.~Lord, M.~Elsabbagh, G.~Baird, and J.~Veenstra-Vanderweele, ``Autism spectrum disorder,'' {\em The lancet}, vol.~392, no.~10146, pp.~508--520, 2018.

\bibitem{orr1998impact}
T.~J. Orr, B.~S. Myles, and J.~K. Carlson, ``The impact of rhythmic entrainment on a person with autism,'' {\em Focus on autism and other developmental disabilities}, vol.~13, no.~3, pp.~163--166, 1998.

\bibitem{wynn2018speech}
C.~J. Wynn, S.~A. Borrie, and T.~P. Sellers, ``Speech rate entrainment in children and adults with and without autism spectrum disorder,'' {\em American Journal of Speech-Language Pathology}, vol.~27, no.~3, pp.~965--974, 2018.

\bibitem{patel2022verbal}
S.~P. Patel, J.~Cole, J.~C. Lau, G.~Fragnito, and M.~Losh, ``Verbal entrainment in autism spectrum disorder and first-degree relatives,'' {\em Scientific reports}, vol.~12, no.~1, p.~11496, 2022.

\bibitem{lahiri2023context}
R.~Lahiri, M.~Nasir, C.~Lord, S.~H. Kim, and S.~Narayanan, ``A context-aware computational approach for measuring vocal entrainment in dyadic conversations,'' in {\em ICASSP 2023-2023 IEEE International Conference on Acoustics, Speech and Signal Processing (ICASSP)}, pp.~1--5, IEEE, 2023.

\bibitem{bone2014psychologist}
D.~Bone, C.-C. Lee, M.~P. Black, M.~E. Williams, S.~Lee, P.~Levitt, and S.~Narayanan, ``The psychologist as an interlocutor in autism spectrum disorder assessment: Insights from a study of spontaneous prosody,'' {\em Journal of Speech, Language, and Hearing Research}, vol.~57, no.~4, pp.~1162--1177, 2014.

\bibitem{ruffman2001social}
T.~Ruffman, W.~Garnham, and P.~Rideout, ``Social understanding in autism: Eye gaze as a measure of core insights,'' {\em The Journal of Child Psychology and Psychiatry and Allied Disciplines}, vol.~42, no.~8, pp.~1083--1094, 2001.

\bibitem{turner1999annotation}
M.~Turner, ``Annotation: Repetitive behaviour in autism: A review of psychological research,'' {\em The Journal of Child Psychology and Psychiatry and Allied Disciplines}, vol.~40, no.~6, pp.~839--849, 1999.

\bibitem{de2020computer}
R.~A.~J. De~Belen, T.~Bednarz, A.~Sowmya, and D.~Del~Favero, ``Computer vision in autism spectrum disorder research: a systematic review of published studies from 2009 to 2019,'' {\em Translational psychiatry}, vol.~10, no.~1, p.~333, 2020.

\bibitem{pandey2020guided}
P.~Pandey, A.~Prathosh, M.~Kohli, and J.~Pritchard, ``Guided weak supervision for action recognition with scarce data to assess skills of children with autism,'' in {\em Proceedings of the AAAI Conference on Artificial Intelligence}, vol.~34, pp.~463--470, 2020.

\bibitem{sabater2021one}
A.~Sabater, L.~Santos, J.~Santos-Victor, A.~Bernardino, L.~Montesano, and A.~C. Murillo, ``One-shot action recognition in challenging therapy scenarios,'' in {\em Proceedings of the IEEE/CVF conference on computer vision and pattern recognition}, pp.~2777--2785, 2021.

\bibitem{loth2018facial}
E.~Loth, L.~Garrido, J.~Ahmad, E.~Watson, A.~Duff, and B.~Duchaine, ``Facial expression recognition as a candidate marker for autism spectrum disorder: how frequent and severe are deficits?,'' {\em Molecular autism}, vol.~9, pp.~1--11, 2018.

\bibitem{zhang2022discriminative}
N.~Zhang, M.~Ruan, S.~Wang, L.~Paul, and X.~Li, ``Discriminative few shot learning of facial dynamics in interview videos for autism trait classification,'' {\em IEEE Transactions on Affective Computing}, vol.~14, no.~2, pp.~1110--1124, 2022.

\bibitem{kojovic2021using}
N.~Kojovic, S.~Natraj, S.~P. Mohanty, T.~Maillart, and M.~Schaer, ``Using 2d video-based pose estimation for automated prediction of autism spectrum disorders in young children,'' {\em Scientific Reports}, vol.~11, no.~1, p.~15069, 2021.

\bibitem{Ali2022VideobasedBU}
A.~Ali, F.~Negin, S.~Th{\"u}mmler, and F.~Br{\'e}mond, ``Video-based behavior understanding of children for objective diagnosis of autism,'' in {\em VISIGRAPP}, 2022.

\bibitem{bommasani2021opportunities}
R.~Bommasani, D.~A. Hudson, E.~Adeli, R.~Altman, S.~Arora, S.~von Arx, M.~S. Bernstein, J.~Bohg, A.~Bosselut, E.~Brunskill, {\em et~al.}, ``On the opportunities and risks of foundation models,'' {\em arXiv preprint arXiv:2108.07258}, 2021.

\bibitem{team2023gemini}
G.~Team, R.~Anil, S.~Borgeaud, Y.~Wu, J.-B. Alayrac, J.~Yu, R.~Soricut, J.~Schalkwyk, A.~M. Dai, A.~Hauth, {\em et~al.}, ``Gemini: a family of highly capable multimodal models,'' {\em arXiv preprint arXiv:2312.11805}, 2023.

\bibitem{Saab2024CapabilitiesOG}
K.~Saab, T.~Tu, W.-H. Weng, R.~Tanno, D.~Stutz, E.~Wulczyn, F.~Zhang, T.~Strother, C.~Park, E.~Vedadi, J.~Z. Chaves, {\em et~al.}, ``Capabilities of gemini models in medicine,'' {\em ArXiv}, vol.~abs/2404.18416, 2024.

\bibitem{Yildirim2024MultimodalHA}
N.~Yildirim, H.~Richardson, M.~T.~A. Wetscherek, J.~Bajwa, J.~Jacob, {\em et~al.}, ``Multimodal healthcare ai: Identifying and designing clinically relevant vision-language applications for radiology,'' {\em Proceedings of the CHI Conference on Human Factors in Computing Systems}, 2024.

\bibitem{Nori2023CapabilitiesOG}
H.~Nori, N.~King, S.~M. McKinney, D.~Carignan, and E.~Horvitz, ``Capabilities of gpt-4 on medical challenge problems,'' {\em ArXiv}, vol.~abs/2303.13375, 2023.

\bibitem{butler2022remote}
L.~K. Butler, C.~La~Valle, S.~Schwartz, J.~B. Palana, C.~Liu, N.~Peterman, L.~Shen, and H.~Tager-Flusberg, ``Remote natural language sampling of parents and children with autism spectrum disorder: Role of activity and language level,'' {\em Frontiers in Communication}, vol.~7, p.~820564, 2022.

\bibitem{lord2012autism}
C.~Lord, M.~Rutter, P.~DiLavore, S.~Risi, K.~Gotham, S.~Bishop, {\em et~al.}, ``Autism diagnostic observation schedule--2nd edition (ados-2),'' {\em Los Angeles, CA: Western Psychological Corporation}, vol.~284, pp.~474--478, 2012.

\bibitem{wang2023domino}
P.~Wang, O.~Golovneva, A.~Aghajanyan, X.~Ren, M.~Chen, A.~Celikyilmaz, and M.~Fazel-Zarandi, ``Domino: A dual-system for multi-step visual language reasoning,'' {\em arXiv preprint arXiv:2310.02804}, 2023.

\bibitem{yang2024empowering}
Y.~Yang, X.~Zhang, J.~Xu, and W.~Han, ``Empowering vision-language models for reasoning ability through large language models,'' in {\em ICASSP 2024-2024 IEEE International Conference on Acoustics, Speech and Signal Processing (ICASSP)}, pp.~10056--10060, IEEE, 2024.

\bibitem{lin2024navcot}
B.~Lin, Y.~Nie, Z.~Wei, J.~Chen, S.~Ma, J.~Han, H.~Xu, X.~Chang, and X.~Liang, ``Navcot: Boosting llm-based vision-and-language navigation via learning disentangled reasoning,'' {\em arXiv preprint arXiv:2403.07376}, 2024.

\bibitem{radford2023robust}
A.~Radford, J.~W. Kim, T.~Xu, G.~Brockman, C.~McLeavey, and I.~Sutskever, ``Robust speech recognition via large-scale weak supervision,'' in {\em International conference on machine learning}, pp.~28492--28518, PMLR, 2023.

\bibitem{achiam2023gpt}
J.~Achiam, S.~Adler, S.~Agarwal, L.~Ahmad, I.~Akkaya, F.~L. Aleman, D.~Almeida, J.~Altenschmidt, S.~Altman, S.~Anadkat, {\em et~al.}, ``Gpt-4 technical report,'' {\em arXiv preprint arXiv:2303.08774}, 2023.

\bibitem{zhang2024llavanextvideo}
Y.~Zhang, B.~Li, h.~Liu, Y.~j. Lee, L.~Gui, D.~Fu, J.~Feng, Z.~Liu, and C.~Li, ``Llava-next: A strong zero-shot video understanding model,'' April 2024.

\bibitem{cheng2024videollama}
Z.~Cheng, S.~Leng, H.~Zhang, Y.~Xin, X.~Li, G.~Chen, Y.~Zhu, W.~Zhang, Z.~Luo, D.~Zhao, {\em et~al.}, ``Videollama 2: Advancing spatial-temporal modeling and audio understanding in video-llms,'' {\em arXiv preprint arXiv:2406.07476}, 2024.

\bibitem{dubey2024llama}
A.~Dubey, A.~Jauhri, A.~Pandey, A.~Kadian, A.~Al-Dahle, A.~Letman, A.~Mathur, A.~Schelten, A.~Yang, A.~Fan, {\em et~al.}, ``The llama 3 herd of models,'' {\em arXiv preprint arXiv:2407.21783}, 2024.

\bibitem{bai2024hallucination}
Z.~Bai, P.~Wang, T.~Xiao, T.~He, Z.~Han, Z.~Zhang, and M.~Z. Shou, ``Hallucination of multimodal large language models: A survey,'' {\em arXiv preprint arXiv:2404.18930}, 2024.

\end{thebibliography}
\bibliographystyle{ieeetr}

\end{document}